\documentclass{article}

\usepackage{PRIMEarxiv}
\usepackage{authblk}

\usepackage[utf8]{inputenc} 
\usepackage[T1]{fontenc}    
\usepackage{hyperref}       
\usepackage{url}            
\usepackage{booktabs}       
\usepackage{amsfonts}       
\usepackage{nicefrac}       
\usepackage{microtype}      
\usepackage{lipsum}
\usepackage{fancyhdr}       
\usepackage{graphicx}       
\graphicspath{{media/}}     
\usepackage{longtable}
\usepackage{caption}
\usepackage{multirow}
\usepackage{amsmath}
\usepackage{float}
\usepackage{pdfpages}
\usepackage{ulem}

\title{
Synthetically Enhanced: Unveiling Synthetic Data’s Potential in Medical Imaging Research
}

\newcommand{\cofirst}{\textsuperscript{*}}
\newcommand{\cosenior}{\textsuperscript{\dag}}

\author[1,2,\cofirst]{Bardia Khosravi, MD MPH MHPE}
\author[3,\cofirst]{Frank Li, PhD}
\author[3]{Theo Dapamede, MD PhD}
\author[1,2]{Pouria Rouzrokh, MD MPH MHPE}
\author[1]{Cooper U. Gamble}
\author[3]{Hari M. Trivedi, MD}
\author[2]{Cody C. Wyles, MD}
\author[4]{Andrew B. Sellergren, BA}
\author[5]{Saptarshi Purkayastha, PhD}
\author[1,\cosenior]{Bradley J. Erickson, MD PhD}
\author[3,\cosenior]{Judy W. Gichoya, MD MS}

\affil[1]{Department of Radiology, Mayo Clinic, Rochester, MN, USA}
\affil[2]{Department of Orthopedic Surgery, Mayo Clinic, Rochester, MN, USA}
\affil[3]{Department of Radiology, Emory University, Atlanta, GA, USA}
\affil[4]{Google Health, Google, Palo Alto, CA, USA}
\affil[5]{School of Informatics and Computing, Indiana University–Purdue University, Indianapolis, IN, USA}

\begin{document}
\maketitle
\vspace{-4em} 
\begin{center}
\cofirst Co-first Author \hspace{1em} \cosenior Co-senior Author

(\href{mailto:bje@mayo.edu}{bje@mayo.edu}, \href{mailto:judywawira@emory.edu}{judywawira@emory.edu})

\vspace{0.5cm}
\noindent{\color{red}\rule{\textwidth}{0.4pt}}
\begin{center}
{\color{red}\large\textbf{The final peer-reviewed version of this paper is now published in \href{https://doi.org/10.1016/j.ebiom.2024.105174}{\uline{Lancet eBioMedicine}}.}}
\end{center}
\noindent{\color{red}\rule{\textwidth}{0.4pt}}

\end{center}
\vspace{2em} 
\begin{abstract}
Chest X-rays (CXR) are the most common medical imaging study and are used to diagnose multiple medical conditions. This study examines the impact of synthetic data supplementation, using diffusion models, on the performance of deep learning (DL) classifiers for CXR analysis. We employed three datasets: CheXpert, MIMIC-CXR, and Emory Chest X-ray, training conditional denoising diffusion probabilistic models (DDPMs) to generate synthetic frontal radiographs. Our approach ensured that synthetic images mirrored the demographic and pathological traits of the original data. Evaluating the classifiers' performance on internal and external datasets revealed that synthetic data supplementation enhances model accuracy, particularly in detecting less prevalent pathologies. Furthermore, models trained on synthetic data alone approached the performance of those trained on real data. This suggests that synthetic data can potentially compensate for real data shortages in training robust DL models. However, despite promising outcomes, the superiority of real data persists.
\end{abstract}


\section{Introduction}
Chest X-rays (CXR) are widely used as the primary imaging modality for a variety of conditions, ranging from acute respiratory distress to chronic pathologies like lung cancer. They are essential for quick and efficient patient triage, especially in emergency settings, and are the most frequently carried out diagnostic imaging exam \cite{National_Health_Services_Government_Statistical_Service_NHS-GSS2023-df}. Deep learning models for CXRs have also recently been used for opportunistic screening of a number of diseases, such as osteoporosis and diabetes mellitus \cite{Jang2022-zv, Pyrros2023-jk}. While chest radiographs are widely used and hold immense potential for diagnostics and screening, their interpretation still requires the expertise of radiologists. The rising demand for these experts and their limited availability creates bottlenecks in healthcare delivery, particularly in underserved areas \cite{Lai2023-zl}. Advances in artificial intelligence (AI) and deep learning (DL) offer a promising avenue to help radiologists by automatically flagging studies with abnormalities.

Several CXR triage tools have been approved by the FDA to detect pathologies like pneumothorax, pleural effusion, and rib fractures \cite{noauthor_2023-ol}. Like other DL-based models, the proposed solutions are not without their shortcomings; most importantly, these models do not always generalize and can suffer from decreased performance when applied to new populations \cite{Yu2022-wa}. For instance, a recent study retrospectively evaluating four commercial tools reported sensitivity ranges of 63\%–90\% and 62\%–95\% across multiple sites for detecting pneumothorax and pleural effusion, respectively \cite{Lind_Plesner2023-mu}. There are several proposed methods for improving generalizability, such as increasing the training sample size and diversity or federated model training \cite{Yang2022-xj,Peng2022-rk}. The former requires combining data from various institutions, but this can be difficult due to concerns about patient privacy \cite{Khosravi2023-uk}. It has also been shown that even after anonymization, CXRs can be re-identified \cite{Macpherson2023-ny,Packhauser2022-wc}.

Generative AI is an emerging research area that aims to develop models that can create realistic content, including text, images, video, and audio, based on a training distribution \cite{OpenAI2023-il}. Image generation models face a trilemma to be an ideal solution—they must produce high-quality images with high diversity in a short period of time—excelling in all three areas simultaneously. Generative adversarial networks (GANs) can quickly create high-quality images but suffer from \textit{mode collapse}, where they cannot generate diverse images even as input prompts change \cite{Bayat2023-wj}. A newer method, denoising diffusion probabilistic models (DDPMs), can create diverse, high quality images and is used in tools such as DALL·E-2 and StableDiffusion \cite{Ramesh2022-en, Podell2023-lg}. Additionally, these models can be conditioned on medically relevant characteristics, creating images with specific attributes \cite{Rouzrokh2022-uy, Sizikova2023-fb, Khosravi2023-vg}. However, as these models denoise an image iteratively, the inference speed can be slow.

With advances in image generation models, there is hope that synthetic data can be used to address some of the aforementioned challenges with model performance and generalizability. Theoretically, by generating high-fidelity synthetic images as a means for dataset augmentation, we can \textit{inject} some distribution characteristics into the training set, improving overall model performance \cite{Ktena2023-bp}. On the other hand, there are some concerns about performance degradation when using synthetic data iteratively, as it may lead to catastrophic interference, or in simpler words, model forgetting \cite{Shumailov2023-zn, Ratcliff1990-kv}. Several studies have used in-domain synthetic data to show a model’s performance will improve when we add model-generated versions of images to their real counterparts, which can lead to better performance in classification scenarios \cite{Frid-Adar2018-re, Chambon2022-dv}. There are also similar studies on other tasks, such as segmentation, showing improved in-domain performance after synthetic augmentation \cite{Pesteie2019-ul, Khosravi2022-pb}. However, these studies did not clearly delineate the underlying causes for improved performance, such as possible leakage of distribution characteristics, increased dataset size, or disentangled distribution. 

The goal of this study is to investigate the effect of synthetic data augmentation in medical imaging research and understand the factors that contribute to model development using a step-by-step methodology. To this end, we first train a conditional DDPM on a subset of the CheXpert dataset and find the optimal hyperparameters for dataset augmentation. Then we create a synthetic replica of this dataset that is up to 10 times larger than the source dataset by creating images with the same demographic and pathologic characteristics as the original dataset. By training several pathology classifiers using a mix of real and synthetic data and testing their performance on internal and external sources, we show the potential and limitations of synthetic data and investigate its modes of failure.

\section{Methods}
\subsection{Dataset Description}
For a comprehensive evaluation of merits of synthetic data in training data expansion, we collected all available frontal chest radiographs from the CheXpert (CXP), MIMIC-CXR (MIMIC) and Emory Chest X-ray (ECXR) datasets \cite{Irvin2019-ta, Johnson2019-ma, Johnson2019-yf,Goldberger2000-ql, Gichoya2022-dl}. All three datasets were annotated using the same automatic natural language processing (NLP) algorithm, CheXpert Labeler \cite{Irvin2019-ta}. CheXpert Labeler categorizes 14 medical conditions into one of four categories based on the radiology reports: 'Present', 'Absent', 'Not Mentioned', and 'Uncertain'. For the purpose of this study, we treated conditions that were 'Not Mentioned' in the reports as being 'Absent' or negative. When training DL models (DDPMs and classifiers), we excluded images that had any 'Uncertain' labels. However, for the testing phase, we included radiographs that had labels marked as 'Uncertain', but we omitted these conditions from our performance metric calculations. The preprocessing pipeline for all images included resizing them to 256 x 256 pixels while preserving the aspect ratio by padding and equalizing the image histogram to 256 bins. We used 4 A100 graphical processing units (GPUs) from NVIDIA (Santa Clara, CA, USA) for all generation and classification experiments. Splitting into train, validation and test sets was done at the patient level for all three datasets.

CXP consists of 161,590 anteroposterior (AP) radiographs from 53,359 unique patients. As our training set (CXP$_{Tr}$), we used 72,053 radiographs from 29,517 individuals who had complete demographic information. Radiographs with 'Uncertain' findings were excluded. For internal testing, we used all images (n=56,448) from the remaining patients (CXP$_{Ts}$). It has been previously shown that deep learning models can have different performance on AP and posteroanterior (PA) images, so we only used PA images and studied the effect of synthetic data supplementation on a dataset of only PA radiographs (MIMIC$_{PA}$; see next) \cite{Lind_Plesner2023-mu}. 

MIMIC-CXR contains 203,456 frontal radiographs from 68,412 distinct patients. We selected 184,587 radiographs from 58,898 individuals as the training set (MIMIC$_{Tr}$). Consistent with CXP curation, radiographs with any 'Uncertain' findings were excluded. For testing (MIMIC$_{Ts}$), all the remaining images (n=5,961) from the remaining patients were used. Although MIMIC$_{Ts}$ is small relative to the other sets, it only served for unbiased model evaluation in one of the experiments (see next). Additionally, we used two splits of the MIMIC dataset, namely MIMIC$_{AP}$ (147,169 AP radiographs from 33,501 patients) and MIMIC$_{PA}$ (96,155 PA radiographs from 45,628 patients), to study the effect of synthetic PA radiographs on AP and PA images from external sources. All dataset splits can be found in our GitHub repository\footnote{\url{https://github.com/BardiaKh/SyntheticallyEnhanced}}. 

Finally, to allow comparison of performance across models, 270,384 frontal radiographs (both AP and PA) from 162,113 patients from the ECXR dataset served as a common test set for all models. 

\subsection{Image Generation}

\begin{figure}
    \centering
    \includegraphics[width=1\linewidth]{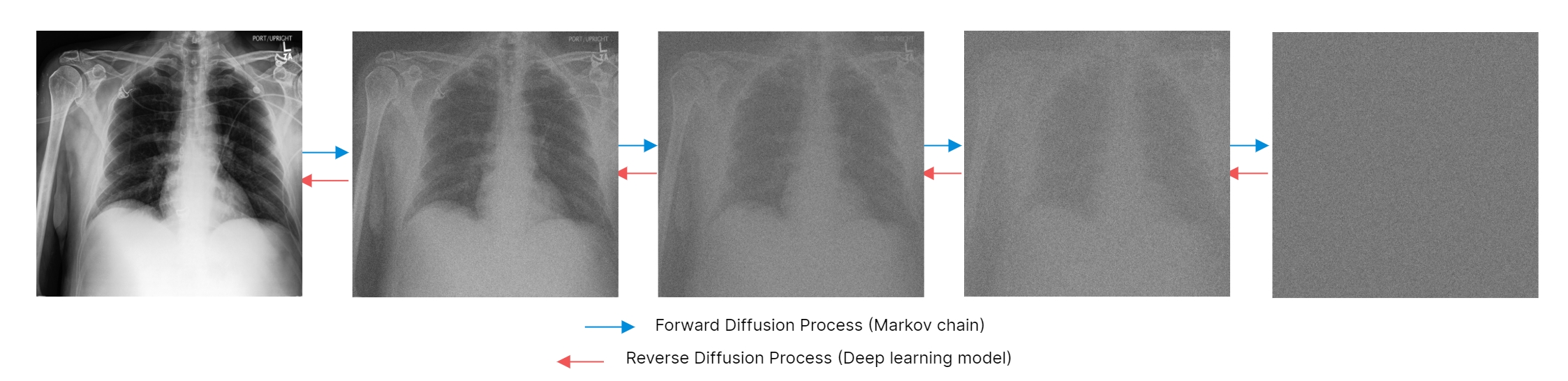}
    \caption{Overview of forward and reverse diffusion processes.}
    \label{fig:fig1}
\end{figure}

We used denoising diffusion probabilistic models (DDPMs) to create synthetic images. DDPMs work by combining forward and reverse diffusion processes. The stepwise incorporation of small amounts of Gaussian noise into a starting image is termed the forward diffusion process. With an increasing number of steps, denoted by total timesteps (\(T\)), the initial image will gradually morph into isotropic Gaussian noise. This sequence is characterized by a Markovian function, which is defined by a noise schedule ($\beta$). This implies that to reach the noisy image at timestep 150 (\(t=150\)), one must sequentially pass through the first 149 timesteps. Nonetheless, due to the characteristics of Gaussian-distributed noise, it is possible to deduce the appearance of an image at step 150 without having to add noise to it in the preceding 149 steps \cite{Ho2020-em}. Using the base image (\(x_0\)), we can determine its noisier counterpart at any chosen timestep (\(x_t\)) with the equation:

\begin{equation}
    x_t = \sqrt{\bar\alpha_t}x_0 + \sqrt{1 - \bar\alpha_t} \epsilon \nonumber
\end{equation}
\begin{align}
    \text{where } \alpha_t = 1 - \beta_t; \quad \bar{\alpha}_t = \prod_{s=0}^{t} \alpha_s \nonumber
\end{align}

The noise schedule ($\beta$) is set in advance of the training, which then establishes the values of \(\alpha_t\) and \(\bar\alpha_t\) (cumulative product from \(\alpha_0\) to \(\alpha_t\) for each timestep \(t\)). In each training step, \(\epsilon\) (from Gaussian distribution) is sampled for each entry in the batch. This noise is paired with the aforementioned equation to produce \(x_t\). The reverse diffusion procedure aims to estimate the noise addition (\(\epsilon\)) between consecutive steps. Contrary to the more direct forward diffusion, achieving reverse diffusion is more demanding and involves training a deep learning (DL) model, often referred to as a diffusion model (see Figure \ref{fig:fig1}). The main goal during training is to reduce the mean squared error (MSE) loss between the noise as predicted by the diffusion model and the actual original noise (\(\epsilon\)), which is precomputed through the forward diffusion technique \cite{Nichol2021-gd}. We employed the Mediffusion package (v0.6.0) to train a generative model based on the CXP$_{Tr}$ set \cite{Khosravi2023-vg}. We used a model with 128 internal channels and a cosine noise schedule with \(T=1000\). We conditioned the model on sex (male / female), age (in decades, e.g., 67 was encoded as 6), race (African American / American Indian / Asian / Pacific Islander / White) and the 14 pathology labels (extracted by CheXpert Labeler). The model was trained for 1.1 million steps. A detailed model configuration can be found in our repository and Table \ref{tab:TE1}. After training, we sampled pure Gaussian noise and denoised it gradually by passing it iteratively through the diffusion model. For faster sampling during inference, we employed implicit sampling with 200 denoising steps \cite{Song2020-gn}. 

To make the generated images correspond to the conditioning variables, we used classifier-free guidance (CFG) \cite{Ho_undated-vz}. Unlike other techniques, CFG does not require the training of a separate classifier to condition the diffusion model. It operates by utilizing a learned \textit{null embedding} that is randomly swapped with actual class embeddings during training. During inference, the CFG scale dictates how closely the generated image corresponds to the conditioning variables. 

Routinely, the CFG scale is set to 4 or 7.5 to generate synthetic images \cite{Ho_undated-vz, Rombach2021-pt}. To investigate the impact of the CFG scale on downstream tasks, we made three replicas of the CXP$_{Tr}$ with the exact same demographic and pathology labels. These synthetic replicas were differentiated by CFG scales set at \{0, 4, 7.5\}. We set aside 10\% of the CXP$_{Tr}$ for validation while using the remaining 90\% to train a classifier aimed at predicting the 14 labels. Through 10 experiments, where we independently substituted the training and validation sets with their synthetic counterparts, we assessed the influence of the CFG scale on classification tasks. The hyperparameters of the pathology classifier are discussed in the next section.

After identifying the most appropriate CFG scale, we generated a large synthetic dataset. In this dataset, every real image in CXP$_{Tr}$ was replicated into 10 synthetic variants. Each variant retained the same demographic and pathology attributes but was created with different initialization seeds to add another level of diversity to the synthetic dataset. 

\subsection{Pathology Classification}
We devised a series of classification tasks to predict 14 condition labels from chest radiographs. A ConvNeXt-base model pretrained on natural image datasets was used for all experiments, hyperparameters presented in Table \ref{tab:TE1} \cite{Wightman2023-mh, Liu2022-fr}. An input size of 256 x 256 pixels was used as it was shown to capture enough information to train a state-of-the-art supervised classifier \cite{Seyyed-Kalantari2021-yf}. Standard online augmentations from the MONAI package (v1.1), including horizontal and vertical flipping, rotation (± 60 degrees), resizing (± 10\%), and translation (± 12 pixels), were employed \cite{The_MONAI_Consortium2020-wr}. We used a learning rate of 0.00001 and a 0.0003 weight decay coupled with the Lion optimizer and binary cross-entropy loss \cite{Chen2023-xx}. To further stabilize the training, we used exponential moving weight averaging (EMA) with a decay factor of 0.9999 \cite{Khosravi2022-kg}. The best model was selected based on the lowest validation loss value.

In order to systematically evaluate the effect of synthetic data on downstream tasks, we defined three sets of experiments. In all experiments, the models have the same hyperparameters, and the only varying factor is the input data. Details and the rationale behind each experiment are as follows:

\subsubsection{Supplementing real data with synthetic data from the same origin:}
The purpose of this experiment was to see if synthetic data from a dataset could increase model performance on the same test set distribution. To this end, we used a graded regimen of synthetic data (100\% to 1000\% in 100\% increments) and added it to the real training set of our classifier. As an example, 300\% supplementation means that we added three times as many synthetically generated sets of images to the original images and used them for training the model. We randomly selected 10\% of the CXP$_{Ts}$ for validation, and the rest were used for model training. Of note, we only used synthetic images for expanding training and did not include synthetic images in the validation set. To have a baseline comparison, we trained a model using only real data (0\% supplementation ratio). We tested the performance of these 11 models on CXP$_{Ts}$, MIMIC-CXR, and ECXR.

\subsubsection{Purely synthetic data:}

This experiment is designed to gauge the performance of a model trained only on synthetic data, simulating an instance where only synthetic data is shared with an outside institution. This experiment helps establish the \textit{utility} of synthetic data alone and shows the extent to which synthetic data can replace real data without sacrificing performance. We used the same split as the previous experiment and excluded any real data from the training set, while keeping the validation set real. Ten models with 10 different quantities of synthetic images (100\% - 1000\% supplementation) were trained and their performance was evaluated on CXP$_{Ts}$, MIMIC-CXR, and ECXR.

\subsubsection{Mixing synthetic data with an external dataset:}

The objective of this experiment was to evaluate the generalizability of our model when trained on a combination of real and synthetic data from different distributions. We used the MIMICTr as the training set (split in 90\%:10\% for training and validation) and mixed 10 different ratios of synthetic data (generated based on CXP$_{Tr}$), similar to previous experiments. These sets of experiments evaluated how different proportions of synthetic data affect the model's generalizability on different datasets. The 10 models were assessed based on their performance on CXP$_{Ts}$, MIMICTs, and ECXR.

\subsection{Evaluation}
We used Fréchet Inception Distance (FID) to evaluate the quality and diversity of generated images \cite{Heusel2017-zp, Lucic2017-yo}. To calculate this, we used images generated with different CFG scales and passed them through an InceptionV3 network that is trained for natural image classification tasks. We then compared the Fréchet distance of the penultimate layer features for real and synthetic images. To assess the performance of pathology classifiers, we used area under receiver operating curve (AUROC) as our main metric. To calculate 95\% confidence intervals (CI), we bootstrapped the results 1000 times and compared models using a paired t-test. Additionally, we adjusted the probability of committing a type I error (\(\alpha\)) using the Bonferroni correction for multiple comparisons. An \(\alpha = 0.05\) was considered the significant level in all instances. Finally, to understand the label distributions, we drew the pathology co-occurrence matrix for each dataset and compared the similarity between them using Pearson's correlation coefficient. Inference speed measurements are reported based on an 80GB A100 GPU. For all statistical analysis, we used the scikit-learn package (v1.3.1) \cite{Pedregosa2012-nf}.

\section{Results}
\subsection{Study Population}
The study utilizes three different datasets with distinct characteristics. The CheXpert dataset is mainly an ambulatory setting dataset and was split into train and testing portions, CXP$_{Tr}$ and CXP$_{Ts}$, respectively. The MIMIC-CXR dataset is an ICU-originated dataset and was mainly used for external validation purposes. Finally, Emory Chest X-ray (ECXR) is a large dataset mixed with ambulatory and ICU cases and was used exclusively for testing. The dataset characteristics are presented in Table \ref{tab:T1} and Table E2.

\newcommand{\includetableone}{}
\ifdefined\includetableone
\renewcommand{\arraystretch}{1.5}
\begin{longtable}{@{}lcccc@{}}
\toprule
\textbf{Variable} &
  \textbf{CXP$_{Tr}$} &
  \textbf{CXP$_{Ts}$} &
  \textbf{MIMIC$\ddag$} &
  \textbf{ECXR} \\* \midrule
\endfirsthead
\multicolumn{5}{c}%
{{\bfseries Table \thetable\ cont.}} \\
\toprule
\textbf{Variable} &
  \textbf{CXP$_{Tr}$} &
  \textbf{CXP$_{Ts}$} &
  \textbf{MIMIC$\ddag$} &
  \textbf{ECXR} \\* \midrule
\endhead
\bottomrule
\endfoot
\endlastfoot
\multicolumn{5}{c}{\textit{\textbf{Dataset Statistics}}} \\
Dataset Origin &
  CA, USA &
  CA, USA &
  MA, USA &
  GA, USA \\
Number of Images &
  72053 &
  56448 &
  243324 &
  270384 \\
Number of Patients &
  29517 &
  23842 &
  63945 &
  162113 \\
\multicolumn{5}{c}{\textit{\textbf{Demographic Information}}} \\
Age (IQR, yrs.) &
  51-75 &
  47-74 &
  52-76 &
  41-67 \\
Sex (female) &
  13583 (46.02\%) &
  \begin{tabular}[c]{@{}c@{}}10847\\ (45.50\%)\end{tabular} &
  \begin{tabular}[c]{@{}c@{}}31610\\ (49.43\%)\end{tabular} &
  \begin{tabular}[c]{@{}c@{}}87777\\ (54.15\%)\end{tabular} \\
\multicolumn{5}{c}{\textit{\textbf{Pathology Labels}}} \\
No Finding &
  \textit{\begin{tabular}[c]{@{}c@{}}pos: 8167\\ (11.33\%)\end{tabular}} &
  \textit{\begin{tabular}[c]{@{}c@{}}pos: 3286\\ (5.82\%)\end{tabular}} &
  \textit{\begin{tabular}[c]{@{}c@{}}pos: 81117\\ (33.34\%)\end{tabular}} &
  \textit{\begin{tabular}[c]{@{}c@{}}pos: 127100\\ (47.01\%)\end{tabular}} \\
Enlarged Cardiomediastinum &
  \textit{\begin{tabular}[c]{@{}c@{}}pos: 3775\\ (5.24\%)\end{tabular}} &
  \textit{\begin{tabular}[c]{@{}c@{}}pos: 2638\\ (4.67\%)\\ unc: 3568\\ (6.32\%)\end{tabular}} &
  \textit{\begin{tabular}[c]{@{}c@{}}pos: 7657\\ (3.15\%)\\ unc: 10001\\ (4.11\%)\end{tabular}} &
  \textit{\begin{tabular}[c]{@{}c@{}}pos: 15475\\ (5.72\%)\end{tabular}} \\
Cardiomegaly &
  \textit{\begin{tabular}[c]{@{}c@{}}pos: 10080\\ (13.99\%)\end{tabular}} &
  \textit{\begin{tabular}[c]{@{}c@{}}pos: 6456\\ (11.44\%)\\ unc: 2656\\ (4.71\%)\end{tabular}} &
  \textit{\begin{tabular}[c]{@{}c@{}}pos: 47673\\ (19.59\%)\\ unc: 6417\\ (2.64\%)\end{tabular}} &
  \textit{\begin{tabular}[c]{@{}c@{}}pos: 49396\\ (18.27\%)\end{tabular}} \\
Lung Lesion &
  \textit{\begin{tabular}[c]{@{}c@{}}pos: 2356\\ (3.27\%)\end{tabular}} &
  \textit{\begin{tabular}[c]{@{}c@{}}pos: 1745\\ (3.09\%)\\ unc: 400\\ (0.71\%)\end{tabular}} &
  \textit{\begin{tabular}[c]{@{}c@{}}pos: 6632\\ (2.73\%)\\ unc: 1192\\ (0.49\%)\end{tabular}} &
  \textit{\begin{tabular}[c]{@{}c@{}}pos: 11191\\ (4.14\%)\end{tabular}} \\
Lung Opacity &
  \textit{\begin{tabular}[c]{@{}c@{}}pos: 32351\\ (44.90\%)\end{tabular}} &
  \textit{\begin{tabular}[c]{@{}c@{}}pos: 30119\\ (53.36\%)\\ unc: 1624\\ (2.88\%)\end{tabular}} &
  \textit{\begin{tabular}[c]{@{}c@{}}pos: 54769\\ (22.51\%)\\ unc: 4023\\ (1.65\%)\end{tabular}} &
  \textit{\begin{tabular}[c]{@{}c@{}}pos: 37380\\ (13.82\%)\end{tabular}} \\
Edema &
  \textit{\begin{tabular}[c]{@{}c@{}}pos: 23430\\ (32.52\%)\end{tabular}} &
  \textit{\begin{tabular}[c]{@{}c@{}}pos: 15893\\ (28.16\%)\\ unc: 5296\\ (9.38\%)\end{tabular}} &
  \textit{\begin{tabular}[c]{@{}c@{}}pos: 29331\\ (12.05\%)\\ Unc:14244\\ (5.85\%)\end{tabular}} &
  \textit{\begin{tabular}[c]{@{}c@{}}pos: 15287\\ (5.65\%)\end{tabular}} \\
Consolidation &
  \textit{\begin{tabular}[c]{@{}c@{}}pos: 4940\\ (6.86\%)\end{tabular}} &
  \textit{\begin{tabular}[c]{@{}c@{}}pos: 3943\\ (6.99\%)\\ unc: 9967\\ (17.66\%)\end{tabular}} &
  \textit{\begin{tabular}[c]{@{}c@{}}pos: 11525\\ (4.74\%)\\ unc: 4598\\ (1.89\%)\end{tabular}} &
  \textit{\begin{tabular}[c]{@{}c@{}}pos: 6331\\ (2.34\%)\end{tabular}} \\
Pneumonia &
  \textit{\begin{tabular}[c]{@{}c@{}}pos: 1599\\ (2.22\%)\end{tabular}} &
  \textit{\begin{tabular}[c]{@{}c@{}}pos: 1195\\ (2.12\%)\\ unc: 6584\\ (11.66\%)\end{tabular}} &
  \textit{\begin{tabular}[c]{@{}c@{}}Pos:17222\\ (7.08\%)\\ unc: 19441 (7.99\%)\end{tabular}} &
  \textit{\begin{tabular}[c]{@{}c@{}}pos: 7303\\ (2.70\%)\end{tabular}} \\
Atelectasis &
  \textit{pos: 13518 (18.76\%)} &
  \textit{\begin{tabular}[c]{@{}c@{}}pos: 9130\\ (16.17\%)\\ unc: 12526\\ (22.19\%)\end{tabular}} &
  \textit{\begin{tabular}[c]{@{}c@{}}pos: 48790\\ (20.05\%)\\ unc: 10965\\ (4.51\%)\end{tabular}} &
  \textit{\begin{tabular}[c]{@{}c@{}}pos: 30815\\ (11.40\%)\end{tabular}} \\
Pneumothorax &
  \textit{\begin{tabular}[c]{@{}c@{}}pos: 9171\\ (12.73\%)\end{tabular}} &
  \textit{\begin{tabular}[c]{@{}c@{}}pos: 4381\\ (7.76\%)\\ unc: 953\\ (1.69\%)\end{tabular}} &
  \textit{\begin{tabular}[c]{@{}c@{}}pos: 11235\\ (4.62\%)\\ unc: 1205\\ (0.50\%)\end{tabular}} &
  \textit{\begin{tabular}[c]{@{}c@{}}pos: 8631\\ (3.19\%)\end{tabular}} \\
Pleural Effusion &
  \textit{\begin{tabular}[c]{@{}c@{}}pos: 32872\\ (45.62\%)\end{tabular}} &
  \textit{\begin{tabular}[c]{@{}c@{}}pos: 21856\\ (38.72\%)\\ unc: 3702\\ (6.56\%)\end{tabular}} &
  \textit{\begin{tabular}[c]{@{}c@{}}pos: 57721\\ (23.72\%)\\ unc: 6202\\ (2.55\%)\end{tabular}} &
  \textit{\begin{tabular}[c]{@{}c@{}}pos: 30205\\ (11.17\%)\end{tabular}} \\
Pleural Other &
  \textit{\begin{tabular}[c]{@{}c@{}}pos: 654\\ (0.91\%)\end{tabular}} &
  \textit{\begin{tabular}[c]{@{}c@{}}pos: 488\\ (0.86\%)\\ unc: 542\\ (0.96\%)\end{tabular}} &
  \textit{\begin{tabular}[c]{@{}c@{}}pos: 2083\\ (0.86\%)\\ unc: 794\\ (0.33\%)\end{tabular}} &
  \textit{\begin{tabular}[c]{@{}c@{}}pos: 4861\\ (1.80\%)\end{tabular}} \\
Fracture &
  \textit{\begin{tabular}[c]{@{}c@{}}pos: 2892\\ (4.01\%)\end{tabular}} &
  \textit{\begin{tabular}[c]{@{}c@{}}pos: 2085\\ (3.69\%)\\ unc: 253\\ (0.45\%)\end{tabular}} &
  \textit{\begin{tabular}[c]{@{}c@{}}pos: 4781\\ (1.96\%)\\ unc: 602\\ (0.25\%)\end{tabular}} &
  \textit{\begin{tabular}[c]{@{}c@{}}pos: 4826\\ (1.78\%)\end{tabular}} \\
Support Devices &
  \textit{\begin{tabular}[c]{@{}c@{}}pos: 45032\\ (62.50\%)\end{tabular}} &
  \textit{\begin{tabular}[c]{@{}c@{}}pos: 33127\\ (58.69\%)\\ unc: 360\\ (0.64\%)\end{tabular}} &
  \textit{\begin{tabular}[c]{@{}c@{}}pos: 73294\\ (30.12\%)\\ unc: 267\\ (0.11\%)\end{tabular}} &
  \textit{\begin{tabular}[c]{@{}c@{}}pos: 119232\\ (44.10\%)\end{tabular}} \\* \bottomrule
\caption{Study population characteristics. Pathology labels extracted using CheXpert labeler from the radiology reports are presented at an image level.
Abbreviations: CXPTr, CheXpert Train; CXPTs, CheXpert Test; ECXR, Emory Chest X-ray; pos, Positive finding by Labeler; unc: Uncertain Finding by Labeler.
‡ Demographic information is only available for 60,523 patients.}
\label{tab:T1}\\
\end{longtable}
\fi

\ifdefined\includetabletwo
\begin{table}[htb]
\centering
\renewcommand{\arraystretch}{1.5}
\begin{tabular}{@{}cc|cccc@{}}
\cmidrule(l){3-6}
\multicolumn{1}{l}{}                                         & \multicolumn{1}{l}{} & \multicolumn{4}{c}{\textbf{Validation set}} \\ \cmidrule(l){3-6} 
\multicolumn{1}{l}{}                                         & \multicolumn{1}{l}{} & Real     & CFG = 0  & CFG = 4  & CFG = 7.5  \\ \cmidrule(l){3-6}
\multicolumn{1}{|c|}{\multirow{4}{*}{\textbf{\rotatebox[origin=c]{90}{Training set}}}} & Real                 & 0.8053   & 0.7911   & 0.9199   & 0.9245     \\
\multicolumn{1}{|c|}{} & CFG = 0   & 0.7969 & 0.8016 & -      & -      \\
\multicolumn{1}{|c|}{} & CFG = 4   & 0.7406 & -      & 0.9671 & -      \\
\multicolumn{1}{|c|}{} & CFG = 7.5 & 0.6983 & -      & -      & 0.9839 \\ \cmidrule(l){2-6} 
\end{tabular}
\caption{The effect of CFG scale of generated images on downstream classifier model performance. All numbers are presented as AUROCs.   
Abbreviations: CFG, classifier-free guidance.}
\label{tab:T2}
\end{table}
\fi

\ifdefined\includetableEone
\begin{table}[H]
\centering
\begin{tabular}{@{}ll@{}}
\toprule
\textbf{Hyperparameter}                                & \multicolumn{1}{c}{\textbf{Value}} \\ \midrule
\multicolumn{2}{c}{\textit{\textbf{Diffusion Model}}}                                                \\
Input Shape                                            & 256 x 256 pixels                   \\
Noise Schedule                                         & Cosine                             \\
Total Timesteps (T)                                    & 1000                               \\
Attention Resolution                                   & 32, 16, 8                          \\
UNet Channels                                          & 128                                \\
UNet Channel Multipliers                               & 1, 1, 2, 2, 4, 4                   \\
Number of Residual Blocks                              & 2                                  \\
Number of Attention Heads                              & 4                                  \\
Number of Conditioning Classes                         & 22 (14 pathologies, 1 age, 2 sexes, 5 races)                                 \\
Guidance Drop Rate (During Training Only)              & 0.1                                \\
\multicolumn{2}{c}{\textit{\textbf{Classifier Models}}}                                              \\
Architecture                                           & ConvNeXt-base                      \\
Pretraining Dataset                                    & ImageNet                           \\
Input Size                                             & 256 x 256 pixels                   \\
Online Augmentations &
  \begin{tabular}[c]{@{}l@{}}- Horizontal and vertical flipping\\ - Rotation (± 60 degrees)\\ - Resizing (± 10\%)\\ - Translation (± 12 pixels)\end{tabular} \\
Learning Rate                                          & 0.00001                            \\
Weight Decay                                           & 0.0003                             \\
Optimizer                                              & Lion                               \\
Loss Function                                          & Binary cross-entropy               \\
Exponential Moving Weight Averaging (EMA) Decay Factor & 0.9999                             \\ \bottomrule
\end{tabular}
\caption{Model hyperparamters.}
\label{tab:TE1}
\end{table}
\fi

\ifdefined\includetableEtwo
\renewcommand{\arraystretch}{1.5}
\begin{longtable}[c]{@{}lcc@{}}
\toprule
\textbf{Variable} &
  \textbf{MIMIC$_{Tr}$\ddag} &
  \textbf{MIMIC$_{Ts}$\dag} \\* \midrule
\endfirsthead
\multicolumn{3}{c}%
{{\bfseries Table \thetable\ cont.}} \\
\toprule
\textbf{Variable} &
  \textbf{MIMIC$_{Tr}$\ddag} &
  \textbf{MIMIC$_{Ts}$\dag} \\* \midrule
\endhead
\bottomrule
\endfoot
\endlastfoot
\multicolumn{3}{c}{\textit{\textbf{Dataset Statistics}}} \\
Dataset Origin &
  MA, USA &
  MA, USA \\
Number of Images &
  184,587 &
  5,961 \\
Number of Patients &
  58,898 &
  4,503 \\
\multicolumn{3}{c}{\textit{\textbf{Demographic Information}}} \\
Age (IQR, yrs.) &
  51-75 &
  47-74 \\
Sex (female) &
  \begin{tabular}[c]{@{}c@{}}29,222\\ (49.61\%)\end{tabular} &
  \begin{tabular}[c]{@{}c@{}}2,124\\ (47.17\%)\end{tabular} \\
\multicolumn{3}{c}{\textit{\textbf{Pathology Labels}}} \\
No Finding &
  \textit{\begin{tabular}[c]{@{}c@{}}pos: 5163\\ (2.80\%)\end{tabular}} &
  \textit{\begin{tabular}[c]{@{}c@{}}pos: 180\\ (3.02\%)\\ unc: 447\\ (7.50\%)\end{tabular}} \\
Enlarged Cardiomediastinum &
  \textit{\begin{tabular}[c]{@{}c@{}}pos: 34570\\ (18.73\%)\end{tabular}} &
  \textit{\begin{tabular}[c]{@{}c@{}}pos: 922\\ (15.47\%)\\ unc: 380\\ (6.37\%)\end{tabular}} \\
Cardiomegaly &
  \textit{\begin{tabular}[c]{@{}c@{}}pos: 4662\\ (2.53\%)\end{tabular}} &
  \textit{\begin{tabular}[c]{@{}c@{}}pos: 225\\ (3.77\%)\\ unc: 224\\ (3.76\%)\end{tabular}} \\
Lung Lesion &
  \textit{\begin{tabular}[c]{@{}c@{}}pos: 27946\\ (15.14\%)\end{tabular}} &
  \textit{\begin{tabular}[c]{@{}c@{}}pos: 2871\\ (48.16\%)\\ unc: 453\\ (7.60\%)\end{tabular}} \\
Lung Opacity &
  \textit{\begin{tabular}[c]{@{}c@{}}pos: 20228\\ (10.96\%)\end{tabular}} &
  \textit{\begin{tabular}[c]{@{}c@{}}pos: 501\\ (8.40\%)\\ unc: 1215\\ (20.38\%)\end{tabular}} \\
Edema &
  \textit{\begin{tabular}[c]{@{}c@{}}pos: 5903\\ (3.20\%)\end{tabular}} &
  \textit{\begin{tabular}[c]{@{}c@{}}pos: 351\\ (5.89\%)\\ unc: 375\\ (6.29\%)\end{tabular}} \\
Consolidation &
  \textit{\begin{tabular}[c]{@{}c@{}}pos: 11723\\ (6.35\%)\end{tabular}} &
  \textit{\begin{tabular}[c]{@{}c@{}}pos: 483\\ (8.10\%)\\ unc: 2554\\ (42.85\%)\end{tabular}} \\
Pneumonia &
  \textit{\begin{tabular}[c]{@{}c@{}}pos: 34175\\ (18.51\%)\end{tabular}} &
  \textit{\begin{tabular}[c]{@{}c@{}}pos: 1027\\ (17.23\%)\\ unc: 1419\\ (23.80\%)\end{tabular}} \\
Atelectasis &
  \textit{\begin{tabular}[c]{@{}c@{}}pos: 9007\\ (4.88\%)\end{tabular}} &
  \textit{\begin{tabular}[c]{@{}c@{}}pos: 44\\ (0.74\%)\\ unc: 77\\ (1.29\%)\end{tabular}} \\
Pneumothorax &
  \textit{\begin{tabular}[c]{@{}c@{}}pos: 38390\\ (20.80\%)\end{tabular}} &
  \textit{\begin{tabular}[c]{@{}c@{}}pos: 906\\ (15.20\%)\\ unc: 671\\ (11.26\%)\end{tabular}} \\
Pleural Effusion &
  \textit{\begin{tabular}[c]{@{}c@{}}pos: 1394\\ (0.76\%)\end{tabular}} &
  \textit{\begin{tabular}[c]{@{}c@{}}pos: 67\\ (1.12\%)\\ unc: 122\\ (2.05\%)\end{tabular}} \\
Pleural Other &
  \textit{\begin{tabular}[c]{@{}c@{}}pos: 3662\\ (1.98\%)\end{tabular}} &
  \textit{\begin{tabular}[c]{@{}c@{}}pos: 134\\ (2.25\%)\\ unc: 257\\ (4.31\%)\end{tabular}} \\
Fracture &
  \textit{\begin{tabular}[c]{@{}c@{}}pos: 52022\\ (28.18\%)\end{tabular}} &
  \textit{\begin{tabular}[c]{@{}c@{}}pos: 717\\ (12.03\%)\\ unc: 33\\ (0.55\%)\end{tabular}} \\
Support Devices &
  \textit{\begin{tabular}[c]{@{}c@{}}pos: 5163\\ (2.80\%)\end{tabular}} &
  \textit{\begin{tabular}[c]{@{}c@{}}pos: 180\\ (3.02\%)\\ unc: 447\\ (7.50\%)\end{tabular}} \\* \bottomrule
\caption{Population characteristics for MIMIC-CXR subsets. Pathology labels extracted using CheXpert labeler from the radiology reports are presented at an image level.
Abbreviations: MIMIC$_{Tr}$, MIMIC Train; MIMIC$_{Ts}$, MIMIIC Test; pos, Positive finding by Labeler; unc: Uncertain finding by Labeler.
\ddag Demographic information is only available for 55,737 patients.
\dag Demographic information is only available for 4,266 patients.}
\label{tab:TE2}\\
\end{longtable}
\fi

\ifdefined\includetableEthree
\setlength{\tabcolsep}{7pt}
\renewcommand{\arraystretch}{1.1}
\footnotesize 
\begin{longtable}[c]{@{}|p{2.5cm}|p{0.5cm}|p{0.5cm}|p{0.5cm}|p{0.5cm}|p{0.5cm}|p{0.5cm}|p{0.5cm}|p{0.5cm}|p{0.5cm}|p{0.5cm}|p{0.5cm}|p{0.5cm}|p{0.5cm}|p{0.5cm}|p{0.5cm}|@{}}
\toprule
\multicolumn{1}{|c|}{\rotatebox{90}{\textbf{Training Data + Supplement}}} &
  \rotatebox{90}{\textbf{Pleural Other}} &
  \rotatebox{90}{\textbf{Fracture}} &
  \rotatebox{90}{\textbf{Support Devices}} &
  \rotatebox{90}{\textbf{Pleural Effusion}} &
  \rotatebox{90}{\textbf{Pneumothorax}} &
  \rotatebox{90}{\textbf{Atelectasis}} &
  \rotatebox{90}{\textbf{Pneumonia}} &
  \rotatebox{90}{\textbf{Consolidation}} &
  \rotatebox{90}{\textbf{Edema}} &
  \rotatebox{90}{\textbf{Lung Lesion}} &
  \rotatebox{90}{\textbf{Lung Opacity}} &
  \rotatebox{90}{\textbf{Cardiomegaly}} &
  \rotatebox{90}{\textbf{Enlarged Cardiomediastinum}} &
  \rotatebox{90}{\textbf{No Finding}} &
  \rotatebox{90}{\textbf{Overall (Macro)}} \\* \midrule
\endfirsthead
\multicolumn{16}{c}%
{{\bfseries Table \thetable\ cont.}} \\
\toprule
\multicolumn{1}{|c|}{\rotatebox{90}{\textbf{Training Data + Supplement}}} &
  \rotatebox{90}{\textbf{Pleural Other}} &
  \rotatebox{90}{\textbf{Fracture}} &
  \rotatebox{90}{\textbf{Support Devices}} &
  \rotatebox{90}{\textbf{Pleural Effusion}} &
  \rotatebox{90}{\textbf{Pneumothorax}} &
  \rotatebox{90}{\textbf{Atelectasis}} &
  \rotatebox{90}{\textbf{Pneumonia}} &
  \rotatebox{90}{\textbf{Consolidation}} &
  \rotatebox{90}{\textbf{Edema}} &
  \rotatebox{90}{\textbf{Lung Lesion}} &
  \rotatebox{90}{\textbf{Lung Opacity}} &
  \rotatebox{90}{\textbf{Cardiomegaly}} &
  \rotatebox{90}{\textbf{Enlarged Cardiomediastinum}} &
  \rotatebox{90}{\textbf{No Finding}} &
  \rotatebox{90}{\textbf{Overall (Macro)}} \\* \midrule
\endhead
\begin{tabular}[c]{@{}l@{}}Real Data Only\\ (baseline)\end{tabular} &
  {\ul 0.768} &
  {\ul 0.793} &
  {\ul 0.862} &
  {\ul 0.891} &
  {\ul 0.879} &
  {\ul 0.692} &
  {\ul 0.785} &
  {\ul 0.759} &
  {\ul 0.853} &
  {\ul 0.821} &
  {\ul 0.728} &
  {\ul 0.882} &
  {\ul 0.702} &
  {\ul 0.861} &
  {\ul 0.805} \\* \midrule
\begin{tabular}[c]{@{}l@{}}Real+Synth\\ Supp: 100\%\end{tabular} &
  0.796 &
  0.828 &
  0.872 &
  0.897 &
  0.895 &
  0.705 &
  0.818 &
  0.764 &
  0.864 &
  0.818 &
  0.735 &
  0.892 &
  0.711 &
  0.867 &
  0.819 \\* \midrule
\begin{tabular}[c]{@{}l@{}}Real+Synth\\ Supp: 200\%\end{tabular} &
  0.797 &
  0.840 &
  0.875 &
  0.897 &
  0.896 &
  0.709 &
  0.828 &
  0.771 &
  0.868 &
  0.836 &
  0.740 &
  0.895 &
  0.724 &
  0.870 &
  0.825 \\* \midrule
\begin{tabular}[c]{@{}l@{}}Real+Synth\\ Supp: 300\%\end{tabular} &
  0.821 &
  0.849 &
  0.876 &
  0.900 &
  0.902 &
  0.709 &
  0.831 &
  0.771 &
  0.869 &
  0.831 &
  0.743 &
  0.898 &
  0.729 &
  0.872 &
  0.829 \\* \midrule
\begin{tabular}[c]{@{}l@{}}Real+Synth\\ Supp: 400\%\end{tabular} &
  0.832 &
  0.855 &
  0.875 &
  0.900 &
  0.906 &
  0.710 &
  0.831 &
  0.775 &
  0.871 &
  0.835 &
  0.744 &
  0.899 &
  0.735 &
  0.871 &
  0.831 \\* \midrule
\begin{tabular}[c]{@{}l@{}}Real+Synth\\ Supp: 500\%\end{tabular} &
  0.832 &
  0.856 &
  0.875 &
  0.901 &
  0.903 &
  0.714 &
  0.839 &
  0.777 &
  0.873 &
  0.836 &
  0.745 &
  0.899 &
  0.735 &
  0.873 &
  0.833 \\* \midrule
\begin{tabular}[c]{@{}l@{}}Real+Synth\\ Supp: 600\%\end{tabular} &
  0.823 &
  0.858 &
  0.875 &
  0.901 &
  0.905 &
  0.715 &
  0.838 &
  0.775 &
  0.874 &
  0.842 &
  0.744 &
  0.900 &
  0.736 &
  0.873 &
  0.833 \\* \midrule
\begin{tabular}[c]{@{}l@{}}Real+Synth\\ Supp: 700\%\end{tabular} &
  \textbf{0.849} &
  0.862 &
  0.876 &
  0.900 &
  0.907 &
  0.714 &
  0.839 &
  0.781 &
  0.873 &
  \textbf{0.847} &
  0.745 &
  0.900 &
  0.745 &
  0.872 &
  0.836 \\* \midrule
\begin{tabular}[c]{@{}l@{}}Real+Synth\\ Supp: 800\%\end{tabular} &
  0.836 &
  0.862 &
  \textbf{0.877} &
  0.900 &
  0.906 &
  \textbf{0.719} &
  0.842 &
  0.781 &
  \textbf{0.876} &
  0.839 &
  \textbf{0.747} &
  \textbf{0.901} &
  \textbf{0.750} &
  \textbf{0.873} &
  0.836 \\* \midrule
\begin{tabular}[c]{@{}l@{}}Real+Synth\\ Supp: 900\%\end{tabular} &
  0.841 &
  0.865 &
  0.875 &
  \textbf{0.902} &
  \textbf{0.910} &
  0.717 &
  0.839 &
  \textbf{0.784} &
  0.875 &
  \textbf{0.847} &
  0.746 &
  \textbf{0.901} &
  0.743 &
  \textbf{0.873} &
  \textbf{0.837} \\* \midrule
\begin{tabular}[c]{@{}l@{}}Real+Synth\\ Supp: 1000\%\end{tabular} &
  0.839 &
  \textbf{0.866} &
  0.876 &
  0.900 &
  0.908 &
  \textbf{0.719} &
  \textbf{0.844} &
  0.781 &
  0.875 &
  \textbf{0.847} &
  0.746 &
  \textbf{0.901} &
  0.745 &
  0.872 &
  \textbf{0.837} \\* \midrule
\begin{tabular}[c]{@{}l@{}}Synth\\ Supp: 100\%\end{tabular} &
  0.744 &
  0.812 &
  0.826 &
  0.882 &
  0.866 &
  0.686 &
  0.785 &
  0.741 &
  0.849 &
  0.784 &
  0.721 &
  0.868 &
  0.693 &
  0.851 &
  0.793 \\* \midrule
\begin{tabular}[c]{@{}l@{}}Synth\\ Supp: 200\%\end{tabular} &
  0.763 &
  0.832 &
  0.849 &
  0.889 &
  0.885 &
  0.693 &
  0.809 &
  0.764 &
  0.861 &
  0.808 &
  0.730 &
  0.884 &
  0.729 &
  0.862 &
  0.811 \\* \midrule
\begin{tabular}[c]{@{}l@{}}Synth\\ Supp: 300\%\end{tabular} &
  0.798 &
  0.841 &
  0.853 &
  0.892 &
  0.891 &
  0.700 &
  0.819 &
  0.762 &
  0.863 &
  0.814 &
  0.733 &
  0.892 &
  0.735 &
  0.862 &
  0.818 \\* \midrule
\begin{tabular}[c]{@{}l@{}}Synth\\ Supp: 400\%\end{tabular} &
  0.814 &
  0.843 &
  0.857 &
  0.894 &
  0.891 &
  0.703 &
  0.825 &
  0.769 &
  0.867 &
  0.822 &
  0.735 &
  0.893 &
  0.738 &
  0.864 &
  0.822 \\* \midrule
\begin{tabular}[c]{@{}l@{}}Synth\\ Supp: 500\%\end{tabular} &
  0.808 &
  0.849 &
  0.858 &
  0.896 &
  0.894 &
  0.703 &
  0.827 &
  0.769 &
  0.867 &
  0.828 &
  0.736 &
  0.893 &
  0.735 &
  0.863 &
  0.823 \\* \midrule
\begin{tabular}[c]{@{}l@{}}Synth\\ Supp: 600\%\end{tabular} &
  0.805 &
  0.849 &
  0.860 &
  0.896 &
  0.897 &
  0.704 &
  0.827 &
  0.769 &
  0.869 &
  0.830 &
  0.734 &
  0.894 &
  0.739 &
  0.864 &
  0.824 \\* \midrule
\begin{tabular}[c]{@{}l@{}}Synth\\ Supp: 700\%\end{tabular} &
  0.838 &
  0.857 &
  0.860 &
  0.895 &
  \textbf{0.899} &
  0.709 &
  0.831 &
  0.775 &
  0.869 &
  \textbf{0.835} &
  0.737 &
  0.893 &
  0.736 &
  0.863 &
  0.828 \\* \midrule
\begin{tabular}[c]{@{}l@{}}Synth\\ Supp: 800\%\end{tabular} &
  0.840 &
  0.857 &
  0.859 &
  0.896 &
  0.898 &
  0.708 &
  0.830 &
  0.772 &
  \textbf{0.870} &
  0.828 &
  0.738 &
  0.895 &
  0.742 &
  \textbf{0.866} &
  0.828 \\* \midrule
\begin{tabular}[c]{@{}l@{}}Synth\\ Supp: 900\%\end{tabular} &
  \textbf{0.849} &
  \textbf{0.860} &
  \textbf{0.861} &
  \textbf{0.897} &
  0.898 &
  \textbf{0.710} &
  \textbf{0.838} &
  0.775 &
  \textbf{0.870} &
  0.834 &
  \textbf{0.739} &
  0.895 &
  0.743 &
  0.864 &
  \textbf{0.831} \\* \midrule
\begin{tabular}[c]{@{}l@{}}Synth\\ Supp: 1000\%\end{tabular} &
  0.839 &
  0.852 &
  \textbf{0.861} &
  \textbf{0.897} &
  \textbf{0.899} &
  0.709 &
  0.832 &
  \textbf{0.777} &
  \textbf{0.870} &
  0.830 &
  0.737 &
  \textbf{0.898} &
  \textbf{0.745} &
  0.864 &
  0.829 \\* \midrule
\begin{tabular}[c]{@{}l@{}}MIMIC\\ (baseline)\end{tabular} &
  {\ul 0.754} &
  {\ul 0.715} &
  {\ul 0.814} &
  {\ul 0.873} &
  {\ul 0.825} &
  {\ul 0.647} &
  {\ul 0.773} &
  {\ul 0.737} &
  {\ul 0.818} &
  {\ul 0.772} &
  {\ul 0.701} &
  {\ul 0.829} &
  {\ul 0.594} &
  {\ul 0.849} &
  {\ul 0.764} \\* \midrule
\begin{tabular}[c]{@{}l@{}}MIMIC + Synth\\ Supp: 100\%\end{tabular} &
  0.798 &
  0.821 &
  0.851 &
  0.888 &
  0.875 &
  0.702 &
  0.812 &
  0.772 &
  0.851 &
  0.804 &
  0.753 &
  0.879 &
  0.716 &
  0.868 &
  0.814 \\* \midrule
\begin{tabular}[c]{@{}l@{}}MIMIC + Synth\\ Supp: 200\%\end{tabular} &
  0.810 &
  0.836 &
  0.855 &
  0.890 &
  0.882 &
  0.708 &
  0.820 &
  0.777 &
  0.856 &
  0.812 &
  0.757 &
  0.884 &
  0.727 &
  \textbf{0.869} &
  0.820 \\* \midrule
\begin{tabular}[c]{@{}l@{}}MIMIC + Synth\\ Supp: 300\%\end{tabular} &
  0.817 &
  0.845 &
  0.857 &
  0.891 &
  0.885 &
  0.712 &
  0.822 &
  0.779 &
  0.858 &
  0.818 &
  0.756 &
  0.886 &
  0.735 &
  0.868 &
  0.824 \\* \midrule
\begin{tabular}[c]{@{}l@{}}MIMIC + Synth\\ Supp: 400\%\end{tabular} &
  0.821 &
  0.846 &
  0.859 &
  0.892 &
  0.887 &
  0.714 &
  0.827 &
  0.781 &
  0.859 &
  0.823 &
  0.758 &
  0.889 &
  0.738 &
  0.868 &
  0.826 \\* \midrule
\begin{tabular}[c]{@{}l@{}}MIMIC + Synth\\ Supp: 500\%\end{tabular} &
  0.829 &
  0.849 &
  0.860 &
  0.893 &
  0.889 &
  0.715 &
  0.827 &
  0.782 &
  0.859 &
  0.824 &
  0.760 &
  0.890 &
  0.740 &
  0.868 &
  0.828 \\* \midrule
\begin{tabular}[c]{@{}l@{}}MIMIC + Synth\\ Supp: 600\%\end{tabular} &
  0.832 &
  0.848 &
  0.860 &
  0.893 &
  0.888 &
  0.715 &
  0.826 &
  0.784 &
  0.860 &
  0.827 &
  0.759 &
  0.890 &
  0.740 &
  \textbf{0.869} &
  0.828 \\* \midrule
\begin{tabular}[c]{@{}l@{}}MIMIC + Synth\\ Supp: 700\%\end{tabular} &
  0.839 &
  0.852 &
  0.861 &
  0.893 &
  0.890 &
  0.717 &
  0.828 &
  0.784 &
  0.861 &
  0.829 &
  0.761 &
  0.892 &
  0.745 &
  \textbf{0.869} &
  0.830 \\* \midrule
\begin{tabular}[c]{@{}l@{}}MIMIC + Synth\\ Supp: 800\%\end{tabular} &
  0.837 &
  0.856 &
  0.862 &
  \textbf{0.894} &
  0.891 &
  0.716 &
  0.829 &
  0.787 &
  0.862 &
  0.831 &
  0.761 &
  0.893 &
  0.747 &
  \textbf{0.869} &
  0.831 \\* \midrule
\begin{tabular}[c]{@{}l@{}}MIMIC + Synth\\ Supp: 900\%\end{tabular} &
  0.838 &
  0.857 &
  0.863 &
  0.893 &
  0.892 &
  0.718 &
  0.833 &
  0.788 &
  \textbf{0.863} &
  \textbf{0.836} &
  0.761 &
  0.893 &
  0.750 &
  0.868 &
  0.833 \\* \midrule
\begin{tabular}[c]{@{}l@{}}MIMIC + Synth\\ Supp: 1000\%\end{tabular} &
  \textbf{0.844} &
  \textbf{0.860} &
  \textbf{0.864} &
  \textbf{0.894} &
  \textbf{0.893} &
  \textbf{0.719} &
  \textbf{0.835} &
  \textbf{0.790} &
  0.862 &
  \textbf{0.836} &
  \textbf{0.762} &
  0.895 &
  \textbf{0.753} &
  \textbf{0.869} &
  \textbf{0.834} \\* \bottomrule
\caption{Detailed classifier model performances on the CheXpert validation set (10\% of CXPTr) that was used for best model selection. All presented numbers are AUROC values. Baseline performance is underlined, and the highest performance in each set of experiments is bolded.}
\label{tab:TabE3}\\
\end{longtable}
\fi
\let\includetableone\undefined

\subsection{Synthetic Data Quality}
\begin{figure}
    \centering
    \includegraphics[width=1\linewidth]{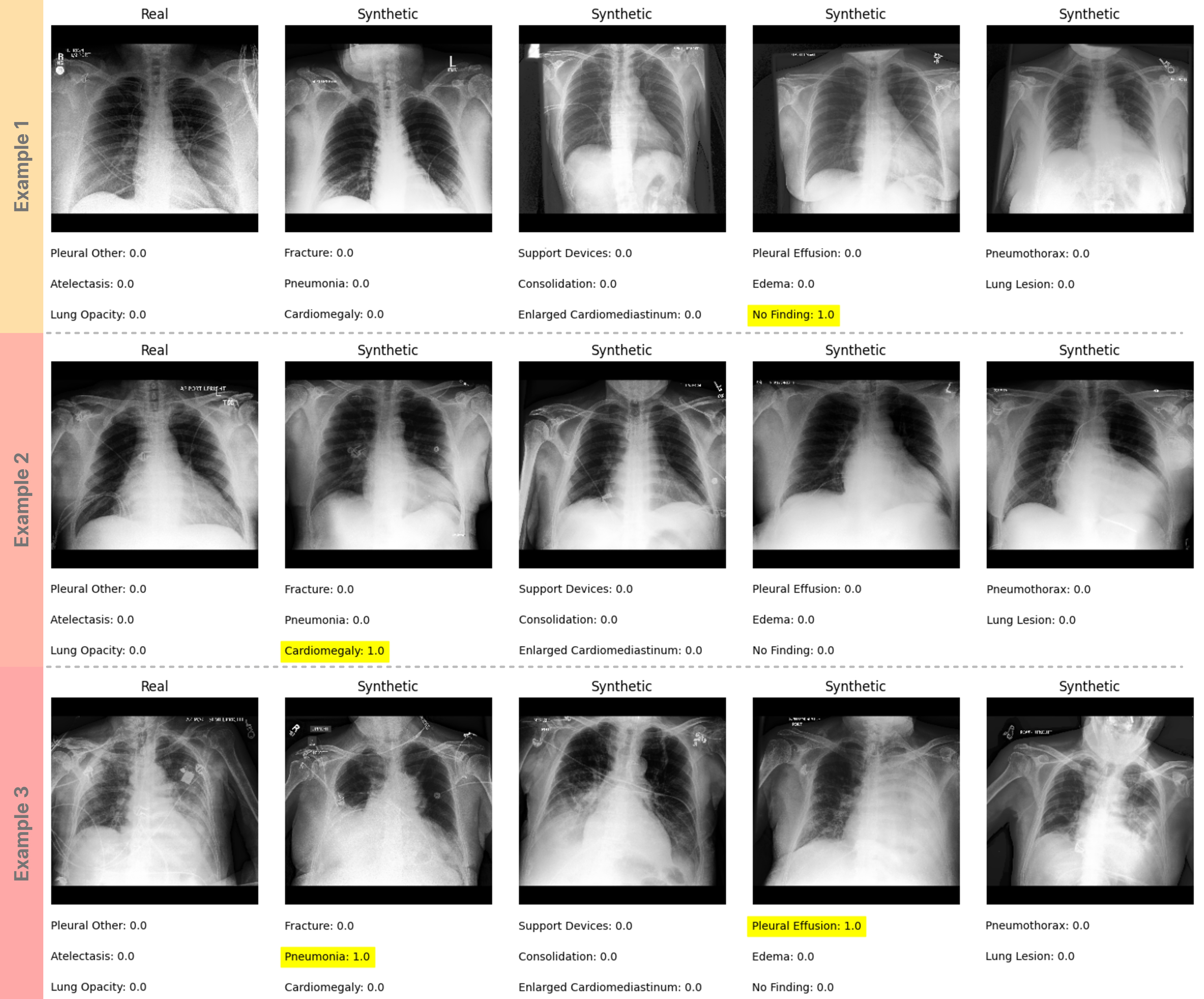}
    \caption{Examples of the real and synthetic images obtained from the diffusion model using different seeds. Presented pathologies are what the model was actually conditioned on.}
    \label{fig:fig2}
\end{figure}

Training the diffusion model took 336 A100 GPU hours, during which the model encountered 40 million sample images. Figure \ref{fig:fig2} represents some of the generated samples along with their conditioned pathology labels. During the initial round of inference, using three different classifier-free guidance (CFG) scales of 0, 4, and 7.5, the model had an FID of 6.4, 7.4, and 13.9, respectively. Additionally, as with CFG scales > 0, two model passes are required to create an image; the inference time for creating a single image with CFG scales of 4 and 7.5 (13.65 seconds and 13.74 seconds, respectively; SD: 0.05) was twice the amount of time for generating images with a CFG scale = 0 (6.84 seconds, SD: 0.04). In terms of performance, the model trained on CFG scale = 0 had an AUROC of 0.797 on the validation set, compared to the model trained on real images (AUROC = 0.805). Table \ref{tab:T2} represents the findings of this set of experiments. Based on these results, we selected a CFG scale of 0 for creating the large synthetic 720K image dataset.

\newcommand{\includetabletwo}{}

\let\includetabletwo\undefined

\subsection{Classification Experiments}
\subsubsection{Supplementing real data with synthetic data from the same origin:}
This experiment was designed to investigate any additive effect that synthetic data might have to improving model performance. By incrementally adding synthetic data to real samples to the classifier training data, AUROC increased from a baseline of 0.782 (no synthetic data) to 0.804 (with 1000\% data supplementation) on CXP$_{Ts}$ (p-value <0.01). Upon external testing on MIMICall adding 1000\% synthetic data supplementation resulted in increasing the model AUROC from 0.749 to 0.770 (p-value <0.01). By analyzing the AP and PA subgroups, we saw an increase in performance on both the MIMICAP and MIMICPA subsets (0.020 and 0.014, respectively, p-value <0.01). External testing on ECXR showed a 0.017 increase in performance, from 0.739 to 0.756, when maximal data supplementation occurred (p-value <0.01). Figure \ref{fig:fig3} summarizes these findings for all the intermediary supplementation steps.

\subsubsection{Purely synthetic data:}
\begin{figure}
    \centering
    \includegraphics[width=1\linewidth]{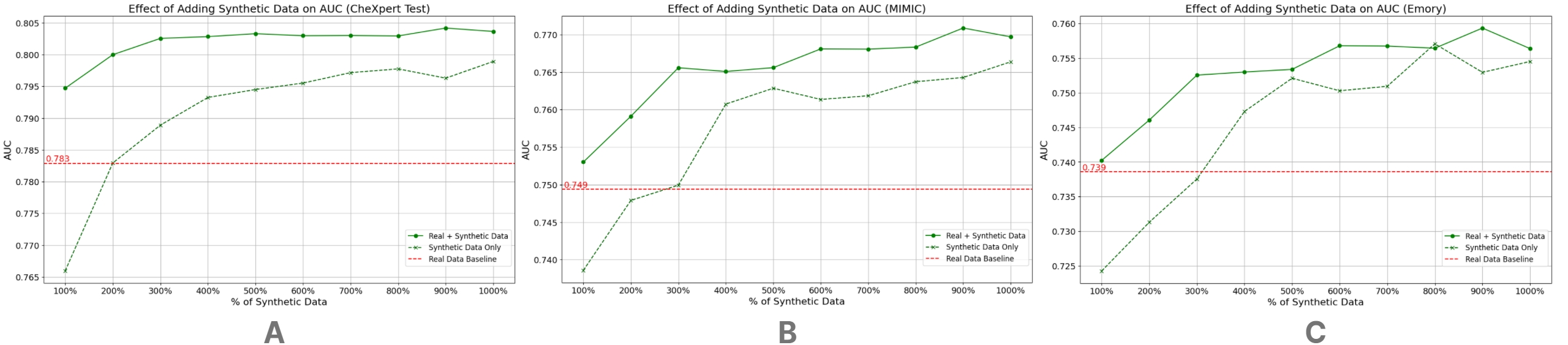}
    \caption{Performance evaluation of models trained on real data, real data supplemented by synthetic data, and synthetic data only on various test sets: (A) CheXpert Test, (B) MIMIC-CXR, and (C) Emory Chest X-ray. The red line in all graphs represents the baseline classifier model’s (trained only on real data from the CheXpert training set) performance on the target dataset.}
    \label{fig:fig3}
\end{figure}

By training the model on synthetic data only, we wanted to explore the hypothetical scenario of sharing synthetic data instead of real data. Our experiments showed similar trends when using synthetic only data to experiments where synthetic and real data were combined. For example, on CXP$_{Ts}$, training the classifier on 200\% of synthetic data showed the same performance as training only on real data (AUCROC of 0.783 in both cases; p-value: 0.98). Similarly, the model achieved similar performance of MIMICall and ECXR when using 300\% synthetic data. However, all models trained on synthetic data alone performed worse than the comparable models trained on real and synthetic data, although the performance gap decreased as the number of synthetic images increased. Figure \ref{fig:fig3} compares the iterative supplementation of synthetic replicas and shows the effect of real data in the three datasets.

\subsubsection{Mixing synthetic data with an external dataset:}
\begin{figure}
    \centering
    \includegraphics[width=0.66\linewidth]{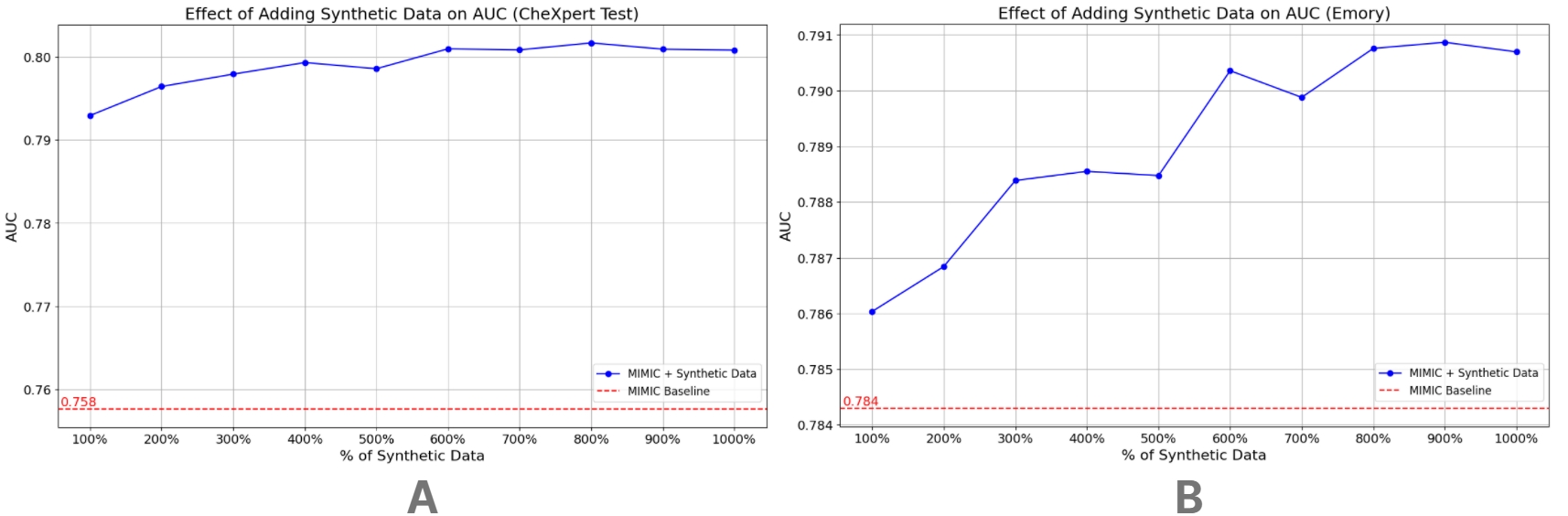}
    \caption{Performance evaluation of models trained on the MIMIC-CXR training set (MIMICTr) with and without supplementation with synthetic data from external sources on various datasets: (A) CheXpert Test, and (B) Emory Chest X-ray.}
    \label{fig:fig4}
\end{figure}

When a classifier model was trained using data from two sites - MIMICTr and 1000\% supplementation of synthetic images from CXP$_{Tr}$, the model performance increased from 0.790 to 0.796 on MIMICTs (p-value <0.01). Additionally, the same model had a 0.007 AUROC increase when tested on ECXR (baseline: 0.784, 1000\% supplementation: 0.791; p-value <0.01). With only 100\% supplementation, the model performance on CXP$_{Ts}$ increased from 0.758 to 0.793  (p-value <0.01). Adding 10 replicas of CXP$_{Tr}$ to MIMICTr, resulted in a final AUROC of 0.801 (p-value <0.01 compared with 100\% supplementation). Figure \ref{fig:fig4} shows the effect of graded synthetic data supplementation on model performance on CXP$_{Ts}$ and ECXR.

\hyperlink{SuppMaterial}{Table E3--E9}, show pathology-specific results for all models. These results show that the largest increase in performance was in pathologies with <5\% prevalence in the population. Additionally, to further investigate the model performance gaps, we created co-occurrence matrices for the three datasets (Figure \ref{fig:figE1}–-\ref{fig:figE3}). We found that MIMIC-CXR and ECXR had the highest correlation (0.862), followed by MIMIC-CXR and CXP (0.846). The least label correlation was between CXP and ECXR, which was 0.789.

\section{Discussion}
Chest radiographs are the most common diagnostic imaging study and are used for the triage or early diagnosis of many medical conditions, making them a prime target for deep learning for automated disease detection. However, there are many reports of performance gaps when the models are tested on various sources. Synthetic data has been used for many years as a way to tackle class imbalances in tabular data analysis \cite{Chawla2002-xe}. However, their use has been limited in medical imaging due to the low quality of the synthetic data. With the emergence of newer techniques, such as diffusion models, there is an opportunity to create high quality diverse medical images and see their effect on downstream task performance. Our study demonstrates that DL models trained on synthetic data can achieve performance levels comparable to those trained on real data, highlighting the viability of synthetic datasets in medical imaging. Additionally, supplementing real datasets with synthetic data significantly enhances model performance and generalizability, particularly in less prevalent pathologies, underscoring the potential of synthetic data in improving model robustness.

Our preliminary experiments showed that a classifier-free guidance (CFG), a hyperparameter determining the adherence of the model to the provided condition, scale of 0 results in synthetic images that are most similar to real images. In other words, it works similarly to the truncation factor ($\Phi$) in generative adversarial networks (GANs). By increasing the CFG scale, we are \textit{overexpressing} the conditioning signal. This makes the model trained on higher CFGs not learn the more nuanced pathology signals in real images, hence the lower performance. Additionally, a model trained on real images has a much easier task of detecting the \textit{overexpressed} signal on synthetic validation sets generated by CFG >0. This idea is also supported by the fact that FID values go up when CFG is increased, which shows that the signals are becoming less diverse.

By supplementing the real radiographs with synthetic replicas of the same origin (both from the CheXpert dataset), the model performance increased, and the increase was incremental with the addition of more synthetic images. We saw a 0.020 increase in average AUROC in internal and external test sets. Interestingly, although the synthetic images were all AP, the performance boost was seen with \textit{both} AP and PA images. This suggests increased dataset diversity is one of the factors contributing to better performance. Additionally, by using synthetic data alone, we found that by using 200-300\% of synthetic data, we can have a model that matches the performance of a model trained on real data. However, the model performance lags behind the model that was trained with real and synthetic data, highlighting the value of real images. 

Finally, by mixing synthetic data with real data from another institution (synthetic CXP$_{Tr}$ and MIMICTr) we observe that the model generalizes better (increased AUROC of 0.043) on the source distribution (CXP$_{Ts}$). However, the increased performance on the CXP validation set which consists of 10\% of CXP$_{Tr}$ that was involved in synthetic image generation, was more pronounced (0.070 AUROC increase) than CXP$_{Ts}$. This suggests that there is information leakage in synthetic data, raising methodological concerns for previous studies in which the test set for evaluating the performance of synthetic data was not explicitly separate from the generative model training or validation source \cite{Frid-Adar2018-re, Chambon2022-dv, Qasim2020-xa}. We believe that to appropriately measure performance gains from synthetic data, the test set must be from a different distribution than the source of the synthetic data; otherwise, performance measurements may be inflated. 

A notable challenge with diffusion models is their slow inference time, necessitated by the iterative denoising process of image generation. Consequently, a single replication of the training set (72,000 images) demands approximately 137 A100 GPU-hours, which can become expensive depending on the compute provider. The decision to balance the enhanced performance derived from synthetic data supplementation against the associated inference costs should be made on a case-by-case basis, taking into consideration the specific requirements and constraints of each research project. Sharing well documented synthetic datasets can also remove the need for individual inference for every model developer. In our study, we found that the most immediate use case of synthetic data is to increase the generalizability of models trained on outside sources, which in many scenarios is worth the investment.

One important implication of using synthetic data is privacy concerns. Several studies have pointed out the possibility of data leaks in generative models. Specifically, it has been shown that if there are multiple copies of a person’s face with the same prompt in a dataset, a diffusion model can associate the face with the person’s name and leak training data \cite{Carlini2023-bg}. This can be concerning in the setting of healthcare, where patient anonymity is of utmost importance. There have been some solutions proposed to ensure training data is not generated, or at least not released, but these are all experimental \cite{Ghalebikesabi2023-gt, Cui2023-if}. As the research frontier tackles synthetic data anonymization, we can only experiment with hypothetical scenarios from publicly available datasets.

Our results should be interpreted with attention to some limitations. First, we used labels extracted by the CheXpert labeler as our conditioning variable for image generation. Since this tool relies on rule-based techniques, there might be abstraction errors that can influence the image quality and ground truth labels for classification \cite{McDermott2020-ai}. However, our results show the lowest FID among the previous works, suggesting proper conditioning \cite{Chambon2022-dv, Weber2023-cp}. Second, we have only used images with one CFG scale. Using a stepwise increase in the CFG scale to create different datasets might help with training a model on both easy and hard cases. This mixture of images with different CFG scales should be further studied. Third, we only investigated the effect of synthetic data supplementation on downstream classification tasks; similar rigorous validation is required for other tasks, such as segmentation and object detection. Finally, we only replicated the dataset without changing the disease prevalence to show only the potential of synthetic data without disentangling disease distribution. In future studies, oversampling of specific pathologies should be investigated \cite{Ktena2023-bp}. Furthermore, exploring online synthetic data generation in the context of active learning presents an intriguing avenue for future research. This approach could allow for more sample-efficient learning, as the model could dynamically generate and learn from data specifically tailored to its current knowledge gaps, potentially making the process both time-efficient and more effective despite the computational cost of generation.

In conclusion, we showed that synthetic data can be useful for training downstream classifier models and that, in large numbers, they can match or outperform the classifiers trained on real data. We further showed the optimal hyperparameters for generating synthetic datasets and showed that the findings are generalizable in two large datasets with more than 500,000 radiographs. Importantly, even a modest amount of synthetic data can close the generalization gap of models trained on other data sources. Finally, we show that real data quality is still superior to synthetic data, and gathering more data should be the first solution for increasing dataset size. 

\bibliographystyle{unsrt}  
\bibliography{references}  

\clearpage
\section*{Supplemental Material}

\renewcommand{\thetable}{E\arabic{table}}
\setcounter{table}{0}
\newcommand{\includetableEone}{}

\let\includetableEone\undefined
\newcommand{\includetableEtwo}{}

\let\includetableEtwo\undefined

\clearpage 
\hypertarget{SuppMaterial}{}
\includepdf[pages=1-16]{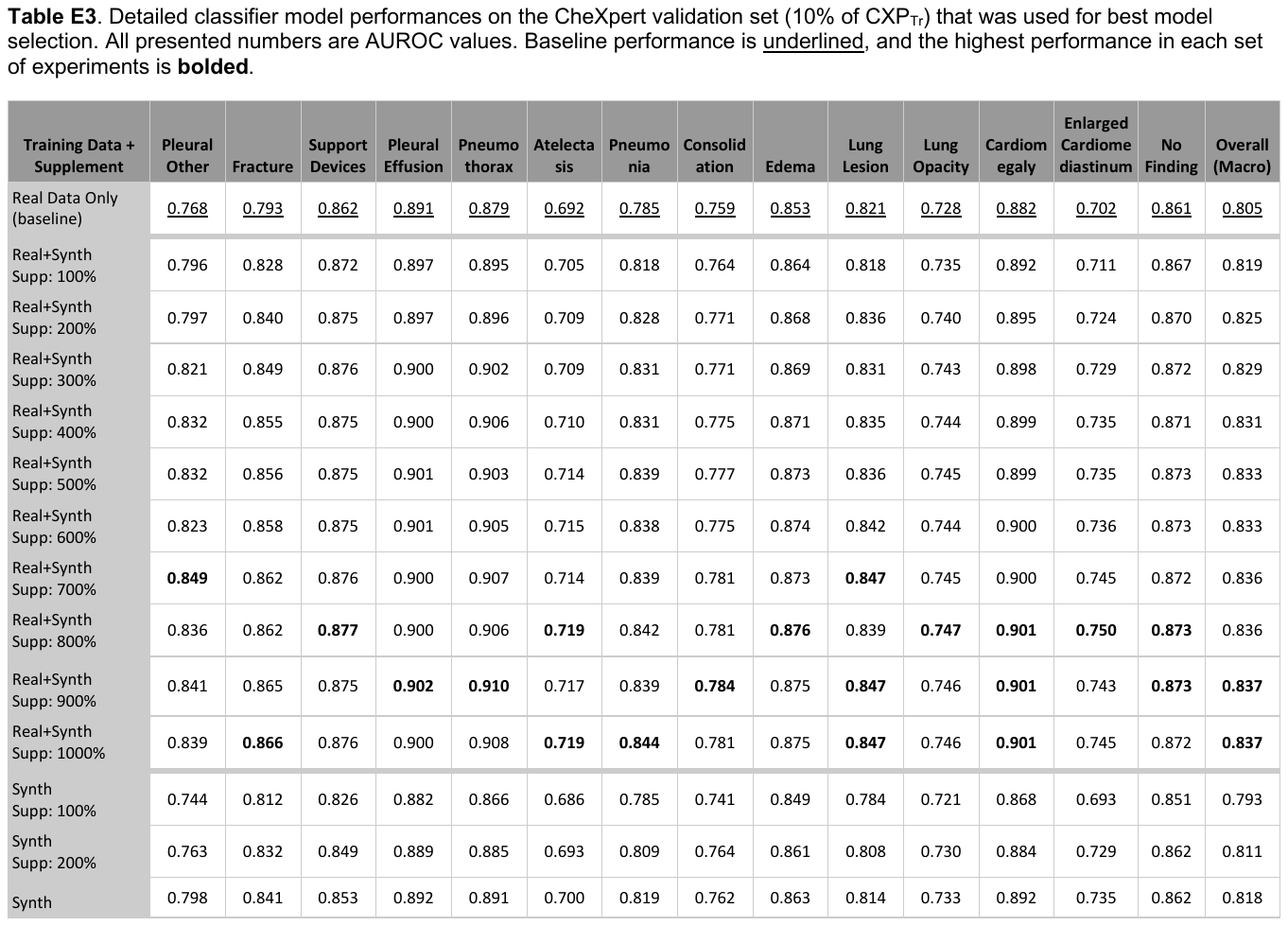}

\renewcommand{\thefigure}{E\arabic{figure}}
\setcounter{figure}{0}
\begin{figure}
    \centering
    \includegraphics[width=1\linewidth]{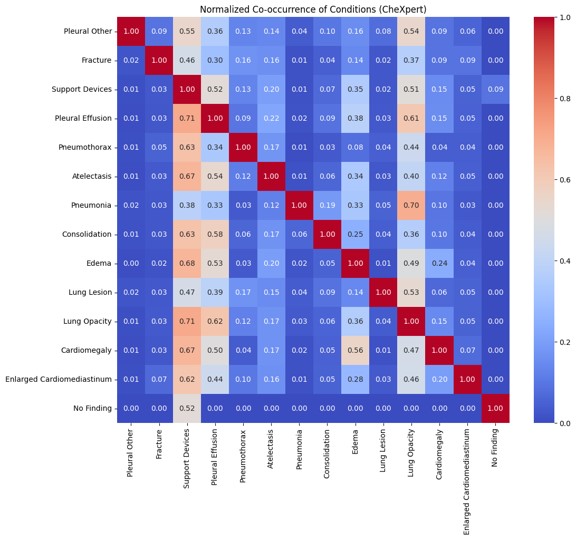}
    \caption{Normalized label co-occurrence matrix for pathologies in the CheXpert dataset. For each condition on the row (\(r\)) of the heatmap, the corresponding column (\(c\)) indicates the ratio of all samples with condition \(r\) that also have condition \(c\).}
    \label{fig:figE1}
\end{figure}
\begin{figure}
    \centering
    \includegraphics[width=1\linewidth]{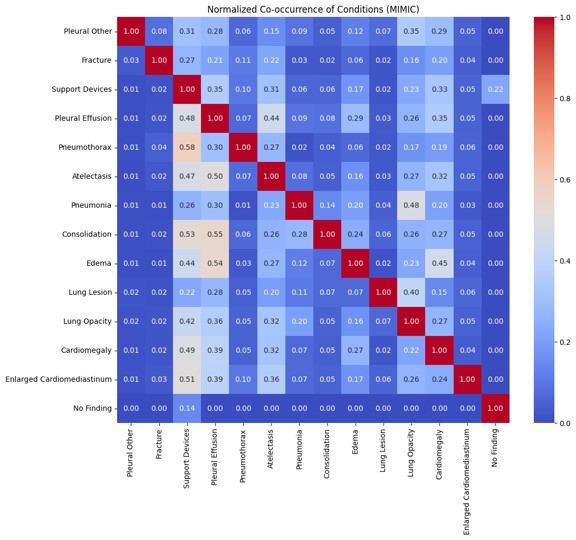}
    \caption{Normalized label co-occurrence matrix for pathologies in the MIMIC-CXR dataset. For each condition on the row (\(r\)) of the heatmap, the corresponding column (\(c\)) indicates the ratio of all samples with condition \(r\) that also have condition \(c\).}
    \label{fig:figE2}
\end{figure}
\begin{figure}
    \centering
    \includegraphics[width=1\linewidth]{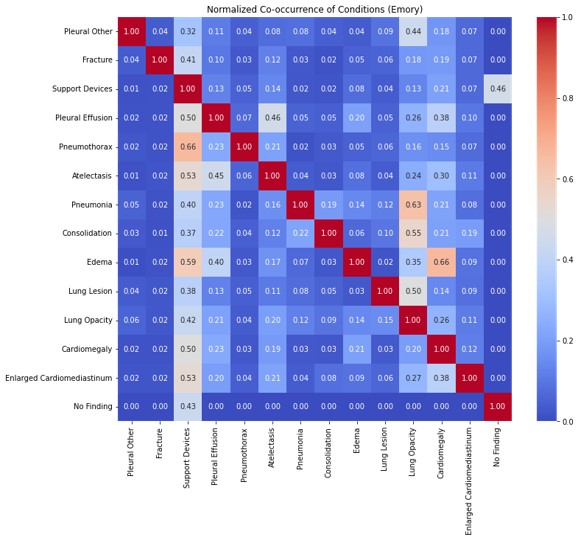}
    \caption{Normalized label co-occurrence matrix for pathologies in the Emory Chest X-ray dataset. For each condition on the row (\(r\)) of the heatmap, the corresponding column (\(c\)) indicates the ratio of all samples with condition \(r\) that also have condition \(c\).}
    \label{fig:figE3}
\end{figure}

\end{document}